\pdfoutput=1

\documentclass[11pt]{article}

\usepackage[]{ACL2023}

\usepackage{times}
\usepackage{latexsym}

\usepackage[T1]{fontenc}

\usepackage[utf8]{inputenc}

\usepackage{microtype}

\usepackage{inconsolata}
\usepackage{xcolor}
\usepackage{hyperref}
\usepackage{lipsum}
\hypersetup{
    colorlinks=true,
    linkcolor=purple,
    filecolor=magenta,      
    urlcolor=purple,
}
\usepackage{url}            
\usepackage{booktabs}       
\usepackage{amsfonts}       
\usepackage{xcolor}         
\usepackage{amssymb}
\usepackage{makecell}
\newcommand{\etal}{\textit{et al}. }

\newcommand{\eg}{\textit{e}.\textit{g}.}

\usepackage{pifont}
\newcommand{\cmark}{\ding{51}}

\usepackage{arydshln}
\usepackage{wrapfig}
\usepackage{multirow}
\usepackage{amsmath}
\usepackage{graphicx}

\usepackage{verbatim}

\title{Where Visual Speech Meets Language: VSP-LLM Framework \\ for Efficient and Context-Aware Visual Speech Processing}

\author{Jeong Hun Yeo$^*$, Seunghee Han$^*$, Minsu Kim, Yong Man Ro$^\dagger$
\\Integrated Vision and Language Lab, KAIST, South Korea\\
\small{\texttt{\{sedne246,gkstmdgml211,ms.k,ymro\}@kaist.ac.kr}}}

\begin{document}
\maketitle
\def\thefootnote{}\footnotetext{$^*$Equal Contribution. $^\dagger$Corresponding Author.}

\begin{abstract}
In visual speech processing, context modeling capability is one of the most important requirements due to the ambiguous nature of lip movements. For example, homophenes, words that share identical lip movements but produce different sounds, can be distinguished by considering the context. In this paper, we propose a novel framework, namely Visual Speech Processing incorporated with LLMs (VSP-LLM), to maximize the context modeling ability by bringing the overwhelming power of LLMs. Specifically, VSP-LLM is designed to perform multi-tasks of visual speech recognition and translation, where the given instructions control the type of task. The input video is mapped to the input latent space of an LLM by employing a self-supervised visual speech model. Focused on the fact that there is redundant information in input frames, we propose a novel deduplication method that reduces the embedded visual features by employing visual speech units. Through the proposed deduplication and Low Rank Adaptation (LoRA), VSP-LLM can be trained in a computationally efficient manner. In the translation dataset, the MuAViC benchmark, we demonstrate that VSP-LLM trained on just 30 hours of labeled data can more effectively translate lip movements compared to the recent model trained with 433 hours of data.
\end{abstract}

\section{Introduction}
Along with audio, visual speech (\eg, lip movements) plays a critical role in human communication. With the increasing acknowledgment of the importance of visual speech, a diverse range of visual-based speech processing technologies~\cite{assael2016lipnet,petridis2016deep,chung2017lip,ma2021towards,ma2022training, yemini2024lipvoicer} is emerging. For instance, Visual Speech Recognition (VSR)~\cite{kim2021cromm,ma2022visual,yeo2023akvsr} allows for the identification of spoken words through the observation of lip movements alone, without the need for audio access. Most recently, the exploration has begun into Visual Speech Translation (VST)~\cite{cheng2023mixspeech}, which directly generates translated text in the target language from the input lip movements of the source language.

One key challenge in visual speech processing is to distinguish homophenes~\cite{kim2022distinguishing}. Homophenes refer to the words having different sounds but showing the same lip movements. Therefore, a crucial aspect of developing visual speech processing systems is in the modeling of context so that the same lip movements can be mapped into correct different pronunciations (that is distinguishing homophenes). Recently, Large~Language~Models~(LLMs) \cite{zhang2022opt,brown2020gpt,workshop2022bloom} are attracting significant attention across various fields \cite{han2023imagebind,wu2023next,fathullah2023prompting}, thanks to their versatility and strong ability to model context. Motivated by the recent success of LLMs, we try to investigate whether the rich context modeling ability of LLMs can be employed in visual speech processing and can mitigate the ambiguity of homophenes, especially focusing on two tasks, VSR and VST.

To this end, in this paper, we propose a new framework named Visual Speech Processing incorporated with LLM (VSP-LLM) that learns the seamless embedding of visual speech into the learned text space of LLMs. VSP-LLM employs a self-supervised visual speech model to embed the input visual speech into phoneme-level representations, where the derived phonetic information can be effectively associated with text \cite{zhang2022speechut}. Moreover, to reduce the computational burden in training along with LLMs, we propose a novel deduplication method that reduces the input sequence lengths of LLMs. Concretely, we employ visual speech units, the discretized representations of the features from a self-supervised model, as indicators for overlapped information between sequences. As the visual speech units can be regarded as pseudo-text \cite{lakhotia-etal-2021-generative}, the visual speech features assigned to the same visual speech units are averaged to reduce the processing of redundant information and improve computational efficiency. Through our analysis, we show that the sequence length can be reduced by approximately 50\% using the proposed deduplication, with minimal performance degradation. Finally, the proposed VSP-LLM is jointly trained to perform VSR and VST with a single model which is the first explored in this paper. We show that by bringing the powerful context modeling ability into visual speech processing, we achieve state-of-the-art performances in both VSR and VST when using the LRS3 \cite{afouras2018lrs3} and MuAViC \cite{anwar2023muavic} datasets as training data. Additionally, our VSP-LLM trained with just 30 hours of data outperforms the recent translation model used 433 hours of training data.

The key contributions of this paper can be summarized as follows: 1) To the best of our knowledge, this is the first work to incorporate visual speech modeling with LLMs and achieve state-of-the-art performances in VSR and VST. 
2) This is the first to work to develop a unified visual speech processing model that can perform both VSR and VST with a single trained model.
3) We propose a novel visual speech deduplication that significantly improves computational efficiency.
4) We show that the proposed VSP-LLM can perform multi-tasks with superior performances even in limited training resource situations, just with 30 hours of labeled data by outperforming the recent translation model.

\section{Related Work}
\subsection{Visual Speech Processing}
Visual speech processing technologies are mainly comprised of two parts, VSR and VST. VSR is a task to recognize the language content by watching lip movements, without any sound. The VSR technologies have greatly progressed with the development of deep learning. Early works \cite{chung2017lrw,stafylakis2017combining,petridis2017lstm,petridis2018end} utilize the CNN  \cite{he2016deep} and the RNN \cite{chung2014empirical, hochreiter1997long} to devise a word-level VSR system. To expand the VSR systems into sentence-level, \cite{chung2017lrs2, afouras2018lrs3} have utilized a multi-stage pipeline to automatically collect large-scale VSR data. Based on the large-scale VSR datasets, researchers \cite{serdyuk2022transformer, ma2021end} have developed the VSR systems from the perspective of architecture, especially the Transformer \cite{vaswani2017attention} have greatly improved the performance of VSR by enabling to capture of the context between any two positions of lip sequences. Moreover, the multimodal learning strategies \cite{zhao2020hearing,afouras2020asr,ren2021learning, ma2021towards,kim2021cromm,kim2022distinguishing,yeo2023multi} have attempted to complement the insufficient visual speech representations by utilizing audio information. A recent self-supervised model known as AV-HuBERT \cite{shi2022learning}, has significantly improved the visual speech representations by predicting the pseudo-label assigned from clustering audio-visual features, with a mask-prediction task like BERT \cite{devlin-etal-2019-bert}. According to the advancement of the VSR system, we can now recognize lip movements quite accurately through state-of-the-art VSR models such as AV-HuBERT. Building upon this, the exploration for VST has begun by introducing a Multilingual Audio-Visual Corpus (MuAViC) \cite{anwar2023muavic} dataset and constructing a VST \cite{cheng2023mixspeech}.

Despite these research efforts, the development of visual speech processing systems enabling multi-task via a unified model, such as VSR and VST, has never been explored in the previous visual speech processing literature.  Hence, the objective of this paper is to develop a unified model to perform multi-tasks, including VSR and VST, by utilizing a rich context modeling ability of LLMs.

\begin{figure*}[t]
	\centering
	\centerline{\includegraphics[width=16cm]{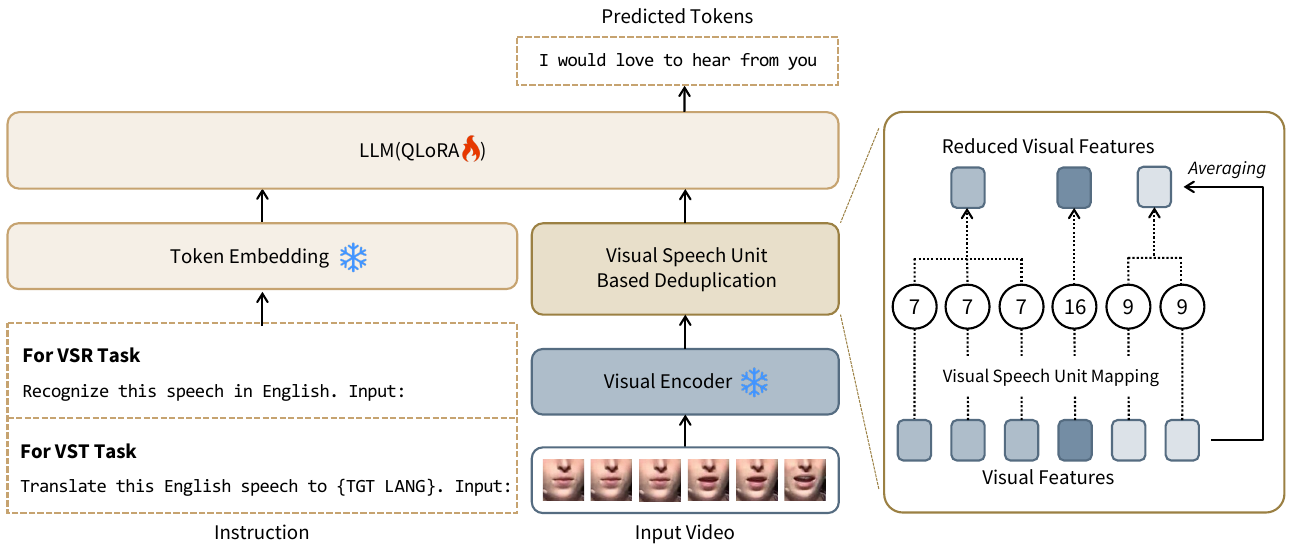}}
    \vspace{-0.3cm}
	\caption{Illustration of our VSP-LLM framework. Visual speech representations encoded from the visual encoder are mapped to visual speech units. Then the visual speech representations are reduced through averaging based on the mapped visual speech units. These reduced representations are fed into the LLM along with text instructions. 
	}
	\label{fig:1}
    \vspace{-0.5cm}
\end{figure*}

\subsection{Integration of speech models and LLMs} 
LLMs have shown remarkable success in various tasks due to their extensive linguistic knowledge and contextual understanding. While leveraging such inherent advantages of LLMs, several studies have tried to seamlessly integrate text-based knowledge with other modalities, particularly in the audio speech domain. For example, AudioPaLM \cite{rubenstein2023audiopalm} has been proposed to build a unified model interacting between text language and audio speech. To naturally bridge the gap between the two modalities, AudioPaLM has developed a multimodal vocabulary composed of discrete tokens representing both text and speech. Fathullah \etal \cite{fathullah2023prompting} have employed LLaMA as a speech recognition decoder so that the speech sequence features obtained from a conformer encoder were designed to be directly mapped into text tokens, the domain of LLaMA. Moreover, Wu \etal \cite{wu2023decoder} have tried to address the inherent problem of mismatched sequence lengths between speech signals and text, while taking LLaMA as a speech translation decoder. So, they have compressed the speech sequence feature and matched its sequence length with that of the text.

However, while the existing studies have primarily focused on incorporating LLMs with the audio speech modality, the exploration of such integration for visual speech processing remains unexplored. In this paper, we propose a novel framework that integrates visual speech processing with LLM. Specifically, we attempt to mitigate the homophenes problem, one of the key challenges in the field of visual speech processing, by leveraging the rich context modeling capabilities of LLM. Additionally, to address the training load issues arising from the integration of the visual speech model and LLM, we introduce the concept of a visual speech unit. Through the implementation of visual speech units, we propose a novel visual speech deduplication method that compresses redundant representations while preserving contextual information.

\section{Method}
\label{sec:3}
Figure \ref{fig:1} shows the overall framework of the proposed Visual Speech Processing incorporated with LLM (VSP-LLM). It includes a visual encoder that embeds the input video into the input space of a pre-trained LLM, a visual speech unit based deduplication module that discards redundant information in contiguous frames, and an instruction embedding component that serves as a task specifier. In the following, we describe each component in detail.

\subsection{Visual-to-Text Space Mapping}
Our primary objective is to employ the rich context modeling capability of LLM in our visual speech modeling. To accomplish this, we need to represent the input video in a manner that aligns closely with linguistic information, thereby facilitating the association between visual inputs and the text space of the pre-trained LLM. Motivated by the recent success of the self-supervised speech models~\cite{hsu2021hubert,shi2022learning} that showed the learned representations are highly correlated with phonetic information (\eg, phoneme) \cite{pasad2023comparative}, we employ AV-HuBERT~\cite{shi2022learning} for our base visual encoder. Then, a learnable visual-to-text embedding layer is introduced to map the visual representations into the input space of LLM. We name this process as visual-to-text space mapping.

To investigate how well the visual representation aligns with the text embedding space of the LLM, we compute the cosine similarity between the visual speech representation and the token embeddings of the LLM, mapping it to the text token with the highest similarity. Figure \ref{fig:2}a shows an example of a textualized visual speech representation. An intriguing observation is that, with well-structured visual-text space mapping, textualized visual speech representations can exhibit pronunciation resembling real words. However, we observe redundant information when mapping entire video frames to text due to the similarity of adjacent frames. For instance, words like 'is' and 'a' are repeated multiple times, and the word 'social' is mapped as a long stretch. This redundancy increases computational load when visual speech representations are fed into LLM. To address this, we propose a novel method called "Visual Speech Unit-based Deduplication" to remove redundancy while retaining semantic content.

\subsection{Visual Speech Unit based Deduplication}
Compared to the length of the input video, the length of the text is much shorter. This is similar to the relationships between speech and text in Automatic Speech Recognition (ASR) \cite{graves2012connectionist}, where the input speech is almost always longer than the output text. Therefore, when we map visual speech representations into text space through visual-to-text space mapping, the resulting embedded output matches the length of the input video frames. If we directly provide it to the LLM, a large computational burden is inevitable. Here, we note that the video is smooth in temporal and the contiguous frames contain overlapped information, and propose to reduce the length of the embedded representation before feeding it to the LLM.

\begin{figure}[t]
	\centering
	\centerline{\includegraphics[width=1.02\linewidth]{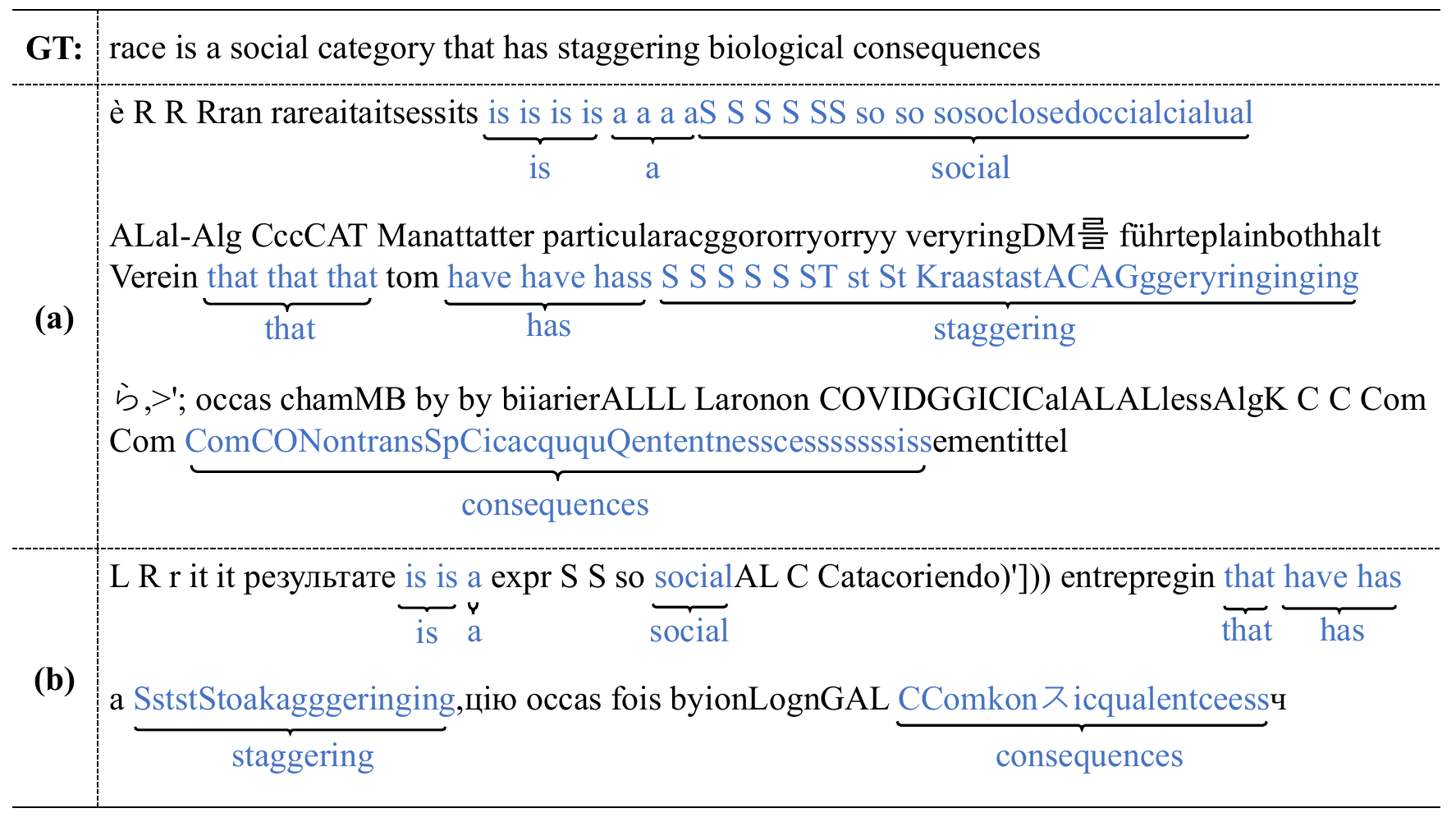}}

	\caption{Textulaization results of the visual speech representations. GT, (a), and (b) indicate the ground truth, textualization without deduplication, and textualization with deduplication, respectively.}
	\label{fig:2}
    \vspace{-0.5cm}
\end{figure}

To this end, we first extract the pronunciation cue from the visual representations through discretization. Recent literature \cite{lakhotia-etal-2021-generative} shows that discretized self-supervised speech features, termed speech units, contain phonetic information while suppressing non-linguistic variations. Motivated by this, we propose to extract a visual version of speech units, namely visual speech units, which can be obtained by performing K-means clustering on the self-supervised visual speech representations. By doing this, we can access the pronunciation information for each video frame without requiring any text input \cite{lee2022textless}. Then, by employing the visual speech units as pseudo text, we investigate the overlapped contiguous frames. Finally, the corresponding visual features are averaged out. For instance, if the obtained visual speech units are $\{7,7,7,16,9,9\}$ as illustrated in Figure \ref{fig:1}, then the visual features at positions 1, 2, and 3 are averaged together, and those at positions 5 and 6 are averaged, resulting in 3 frames. We find that the proposed visual speech unit based deduplication reduces the sequence lengths by about 46.62\% compared to the input video lengths. Most importantly, we observed that the deduplication process does not result in any drop in performance. The reduced visual features, when converted into text (Figure \ref{fig:2}b), maintain the meaning of each word while the duplication of each word has been removed. For instance, the recurrence of 'is' and 'a', which appeared multiple times in the original feature,  is reduced, and the length of 'social', which has a long stretch, is also drastically reduced.

\subsection{Multi-task Learning with Instruction}
One advantage of bridging LLMs into visual speech processing is that we can leverage the versatility of LLMs as well. To investigate this, we train the proposed VSP-LLM with two tasks, VSR and VST. VSR aims to recognize the input silent speech while VST aims not only to predict the recognized speech but also to translate it into the target language. We design the system so that tasks can be controlled by inputting instructions directly into the LLM. When performing the VSR task the instruction is set to as below,

{\small
\begin{verbatim}
Recognize this speech in English. 
Input: ${Dedupped_Visual_Feature}
\end{verbatim}
}

where the deduplicated visual features are inserted after the instruction.
Otherwise, to perform VST, the following instruction is employed.

{\small
\begin{verbatim}
Translate this English speech to ${TGT LANG}. 
Input: ${Dedupped_Visual_Feature}
\end{verbatim}
}

where the target language is used for the position of \texttt{TGT LANG}. The objective function for each task can be written as follows,
\begin{align}
\mathcal{L} = - \sum^{L}_{l=1} \ \log p(y^l|X,I,y^{<l}),
\end{align}
where $X$ is input video, $I$ is instruction used, $y^{l}$ is the $l$-th text token of the ground truth sentence, $y^{<l}$ is the previous predictions, and $L$ is the length of ground truth. Please note that this is the first work exploring a unified framework of VSR and VST. For training, we employ the recently proposed QLoRA~\cite{dettmers2023qlora} to further relieve the computational load in training LLM.

\section{Experiment}
\label{sec:4}
\subsection{Dataset}
 
{\bf Lip Reading Sentences 3 (LRS3)} \cite{afouras2018lrs3} is the most widely-used dataset for VSR, which comprises 433 hours of English audio-visual speech corpus with transcription data. These corpora are collected from the TED and TEDx talks. We utilize the LRS3 dataset to measure the VSR performance of the proposed unified model.

\noindent{\bf Multilingual Audio-Visual Corpus (MuAViC)}  \cite{anwar2023muavic} is a multilingual audio-visual dataset designed for speech recognition and speech-to-text translation. It includes 1200 hours of audio-visual corpus in 9 languages, providing full transcriptions and covering 6 English-to-X translations, as well as 6 X-to-English translation directions. To evaluate the VST performance of our model, we utilize English-to-X translation data from MuAViC dataset, where X can be among four languages, Spanish (Es), French (Fr), Portuguese (Pt), and Italian (It).
 For training our model, we combine the LRS3 dataset and English-to-X translation data of MuAViC.

\begin{table*}[t]
  \renewcommand{\arraystretch}{1.5}
  \renewcommand{\tabcolsep}{2.5mm}
  \centering
  \resizebox{0.95\linewidth}{!}{
  \begin{tabular}{ccccccc}
    \Xhline{3\arrayrulewidth}
    \multicolumn{2}{c}{Method} & \makecell{\textbf{Pre-training} \\ \textbf{Data (hrs)}} & \makecell{\textbf{Labeled} \\ \textbf{Training Data (hrs)}} & \makecell{\textbf{Recognition} \\ \textbf{Task} } & \makecell{\textbf{Translation} \\ \textbf{Task}} & \textbf{WER(\%)}  \\ \hline

    \multirow{8}{*}{\textbf{Supervised}} & 
    \citet{afouras2018lrs3} & - & 1,519 & \cmark & & 58.9  \\
    &\citet{DBLP:conf/interspeech/ShillingfordAHP19} & - & 3,886 & \cmark & & 55.1  \\
    &\citet{makino2019recurrent} & - & 31,000 & \cmark & & 33.6  \\
    &\citet{prajwal2022sub} & - & 2,676 & \cmark & & 30.7  \\
    &\citet{ma2021end} & - & 595 & \cmark &  & 30.4 \\
    &\citet{ma2023auto} & - & 3,448 & \cmark & & 19.1  \\
    &\citet{serdyuk2022transformer} & - & 90,000 & \cmark & & 17.0  \\
    &\citet{chang2023conformers} & - & 100,000 & \cmark & & 12.8  \\
    \hline

    \multirow{11}{*}{\textbf{Self-supervised}} & 
    AV-HuBERT \cite{shi2022learning} & 1,759 & 30 & \cmark &  & 32.5  \\
    &VATLM \cite{zhu2023vatlm}& 1,759 & 30 & \cmark &  & 31.6  \\ 
    &RAVen \cite{haliassos2022jointly}& 1,759 & 30 & \cmark &  & 32.5 \\ 
    &AKVSR \cite{yeo2023akvsr}& 1,759 & 30 & \cmark &  & 29.1 \\ 
    \cdashline{2-7}
    &\textbf{VSP-LLM} & 1,759 & 30 &  \cmark & \cmark & 29.8 \\
    \cline{2-7}
    &AV-HuBERT \cite{shi2022learning}& 1,759 & 433 & \cmark &  & 28.6  \\ 
    &VATLM \cite{zhu2023vatlm}& 1,759 & 433 & \cmark &  & 28.4  \\ 
    &RAVen \cite{haliassos2022jointly}& 1,759 & 433 & \cmark &  & 27.8 \\
    &AKVSR \cite{yeo2023akvsr}& 1,759 & 433 & \cmark &  & 27.6 \\  
    \cdashline{2-7}
    & \textbf{VSP-LLM} & 1,759 & 433 &  \cmark & \cmark  & 26.7  \\
    & \textbf{VSP-LLM(FT)} & 1,759 & 433 &  \cmark & \cmark & 25.4  \\ 
    \Xhline{3\arrayrulewidth}
  \end{tabular}}
  \caption{The performance comparisons with state-of-the-art VSR methods. Compared to the self-supervised methods, the proposed VSP-LLM, which can perform both VSR and VST, achieves state-of-the-art recognition performances. We also evaluate the performance of a fine-tuned VSP-LLM(FT) with an unfrozen visual encoder.}
  \vspace{-0.3cm}
  \label{table:1}
\end{table*}

\begin{table}[t]
\renewcommand{\arraystretch}{1.6}
\renewcommand{\tabcolsep}{2.0mm}
  \centering
  
  \resizebox{0.999\linewidth}{!}{
  \begin{tabular}{ccccccc}
    \Xhline{3\arrayrulewidth}
    \multirow{2}{*}{\textbf{Method}} &  \multirow{2}{*}{\makecell{\textbf{Labeled} \\ \textbf{data(hrs)}}} & \multicolumn{5}{c}{\textbf{BLEU  $\uparrow$}}  \\
    \cline{3-7}
     &  & \textbf{En-It} & \textbf{En-Fr} & \textbf{En-Pt} & \textbf{En-Es} & \textbf{Avg} \\
    \hline
    \citet{anwar2023muavic} & 433 & 15.1 & 16.8 & 15.1 & 19.2 & 16.6 \\ 
    \makecell{AV-HuBERT} & 433 & 16.6 & 19.4 & 17.4 & 21.7 & 18.8 \\
    \makecell{\makecell{Cascaded (AV-HuBERT + MT)}} & 433 & 17.6 & 19.5 & 17.4 & 22.4 & 19.2 \\ 
    \hdashline
    \textbf{VSP-LLM} & 30 & 16.1 & 19.3 & 16.6 & 20.7 & 18.2 \\ 
    \textbf{VSP-LLM} & 433 & \textbf{17.9} & \textbf{22.3} & 18.7 & \textbf{22.7} & \textbf{20.4} \\ 
    \textbf{VSP-LLM(FT)} & 433 & 17.7 & 22.2 & \textbf{19.4} & 22.4 & \textbf{20.4} \\ 
    \Xhline{3\arrayrulewidth}
  \end{tabular}}
  \caption{Experimental results for English to target language (En-X) translation on the MuAViC benchmark.}
  \vspace{-0.6cm}
  \label{table:2}
\end{table}

\subsection{Implementation Details}
\textbf{Preprocessing.} The video is resampled at 25 fps, and facial landmarks are detected using RetinaFace \cite{deng2020retinaface}. Mouth regions are cropped using bounding boxes of size $96 \times 96$ and converted to grayscale. During training, we apply data augmentation by randomly cropping the video to $88 \times 88$ and horizontally flipping it.

\noindent \textbf{Architecture.} We use the AV-HuBERT \textit{large} \cite{shi2022learning} pre-trained on LRS3 \cite{afouras2018lrs3} and VoxCeleb2 English \cite{chung2018voxceleb2} as our visual encoder. In all experiments, except the ablation part, we utilize 200 clustered visual speech units. For the LLM, we adopt LLaMA2-7B \cite{touvron2023llama} and fine-tune it using QLoRA \cite{dettmers2023qlora} with the rank value of 16 and a dropout rate of 5\%. To align the dimensions of the visual representation from the visual encoder to the LLaMA input embedding, we use a single linear layer as our visual-to-text embedding layer. 

\noindent \textbf{Training and evaluation.} We follow AV-HuBERT \cite{ren2021learning} except for the number of updates and learning rate. We conduct training with a learning rate of $5e^{-4}$ and the number of updates is 15K updates for LRS3 1h, 5h, 10h, and 30K updates for LRS3 30h and 433h. For VSP-LLM (FT), the visual encoder is frozen for the first 18K steps and then unfrozen afterward. Adam optimizer is employed for training with $\beta_1 = 0.9$ and $\beta_2 = 0.98$, utilizing a tri-stage learning rate scheduler. The training process is executed on 8 3090 RTX GPUs. For decoding, we use a beam search with a beam width of 20 and a length penalty of 0. We assess the performance of our model using Word Error Rate (WER) for the VSR task and BLEU score \cite{papineni2002bleu} for the VST task. We use total FLOPs per epoch as a metric to measure the model operation count during training.

\subsection{Experimental Results}
\subsubsection{Comparison with State-of-the-arts}
In this subsection, we compare the proposed unified model with state-of-the-art VSR and VST methods. Please note that the proposed model can perform multi-tasks VSR and VST with a single trained model while the other models need a single model per specific task. 

Table \ref{table:1} presents the performance comparisons of the proposed method with state-of-the-art VSR methods on the LRS3 dataset. The top section of Table \ref{table:1} outlines the performance of current supervised approaches that depend on extensive labeled training data, while the lower section presents a comparison with other self-supervised methods. Table \ref{table:1} demonstrates that our approach achieves performance on par with others by employing just 30 hours of labeled data, despite the proposed unified model's ability to handle multiple tasks—VSR and VST—simultaneously. When employing 433 hours of training data, our method achieves a WER of 26.7\%. By fine-tuning the VSP-LLM(FT) with an unfrozen visual encoder, we further enhance our performance, achieving a WER of 25.4\%, surpassing other self-supervised approaches. Moreover, Table \ref{table:1}'s upper part shows that the existing supervised methods record exceptional performance using (tens of) thousands of labeled data. However, it is important to highlight that the proposed unified model can obtain comparable performances to several supervised methods.

Table \ref{table:2} presents the comparison results of VST performance. We construct two baseline models for comparison. The first, AV-HuBERT, is trained similarly to our approach, utilizing both VSR and VST datasets. The second model is a cascaded system that incorporates a pre-trained AV-HuBERT for VSR with a neural machine translation model \cite{fan2021beyond}. Through this comparison, our proposed VSP-LLM demonstrates superior VST performance across four English-to-X translation tasks, achieving BLEU scores of 17.9, 22.3, 18.7, and 22.7 for English to Italian, French, Portuguese, and Spanish, respectively. The VSP-LLM(FT) shows a better performance 19.4 BLUE score on translation from English to Portuguese and comparable performances in other languages. Moreover, it is worth noting that the proposed method achieves an 18.2 BLEU score on average with only 30 hours of labeled data, outperforming the bilingual speech translation model \cite{anwar2023muavic} trained with 433 hours of labeled data.

\begin{figure*}[t]
	\centering
	\centerline{\includegraphics[width=16cm]{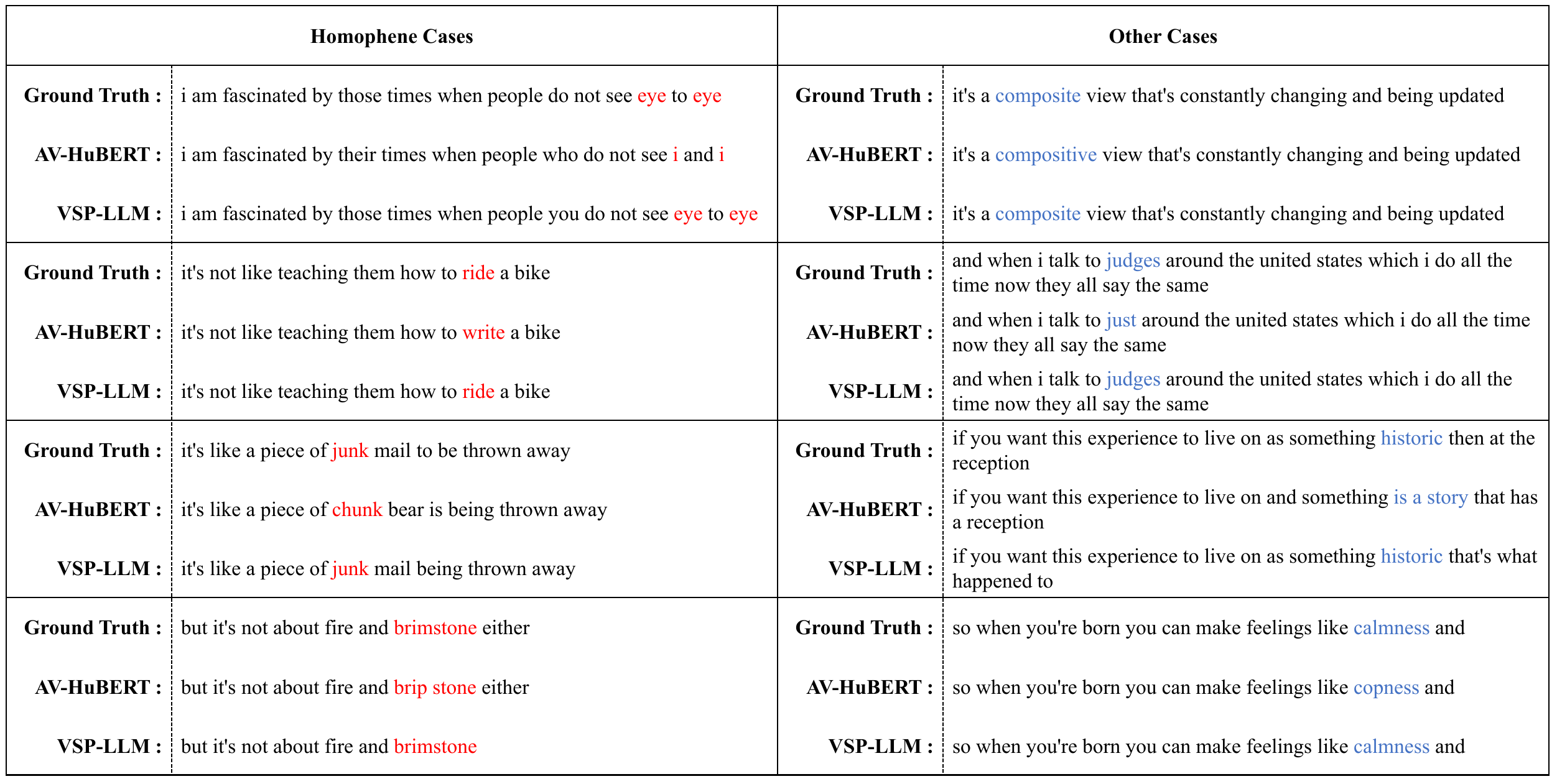}}
        \caption{The qualitative results showing that the contextual modeling ability of LLM, which is adopted in our method, can improve the homophene problem and other confusing cases. The \textcolor[rgb]{0.9,0,0}{red} and \textcolor[rgb]{0,0.2,0.90}{blue} words indicate the wrong predictions from AV-HuBERT. However, as shown in the examples, the proposed method can generate correct words by considering the entire context (\textit{e.g.,} `\textit{i}' to `\textit{eye}').}
        \vspace{-0.5cm}
	\label{fig:3}
\end{figure*}

\begin{figure}[t]
	\centering
	\centerline{\includegraphics[width=0.999\linewidth]{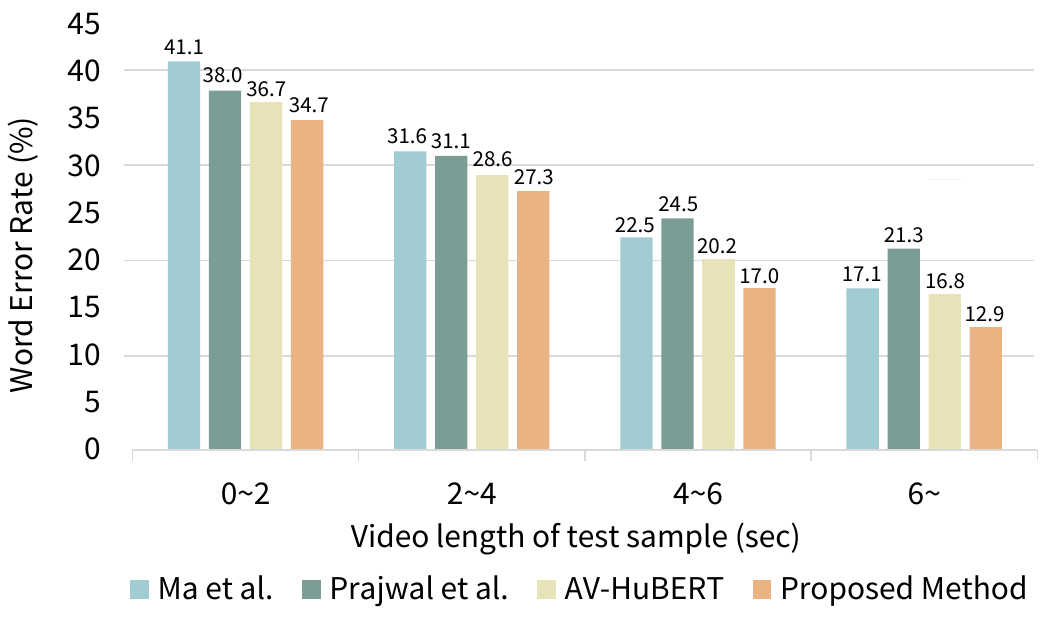}}
        \vspace{-0.1cm}
        \caption{VSR performance analysis on LRS3 with varying video length of test samples. Due to the strength of contextual understanding ability of LLM, the proposed method shows superior performance with longer videos.}
	\label{fig:4}
    \vspace{-0.5cm}
\end{figure}

\subsubsection{Effectiveness of Rich Context Modeling}
\label{sec:rich_context}

We have developed a unified model incorporating LLMs to leverage their advanced context modeling capabilities. Therefore, in this section, we conduct a qualitative experiment to demonstrate the effectiveness of the proposed VSP-LLM in handling homophenes, a challenging problem that requires substantial context understanding to accurately identify homophenes. Figure \ref{fig:3} shows several transcription examples obtained from AV-HuBERT and our model, illustrating how our proposed method accurately generates words by considering the entire context of a sentence. For instance, in a homophene case, AV-HuBERT incorrectly transcribes "i", a word which visually resembles "eye" on the lips, but differs in meaning. On the other hand, our method correctly generates "eye", successfully completing the idiom "eye to eye" to describe mutual understanding between individuals. Similarly, AV-HuBERT's transcription of "write" is contextually inappropriate for a sentence discussing teaching the physical skill of riding a bike. Our method, however, accurately outputs "ride" resulting in the correct phrase "ride a bike". Also, we can observe similar results in the other cases, not the homophene problem only. For example, the proposed method can generate the word “composite” according to standard English usage, unlike AV-HuBERT, which erroneously outputs "compositive". These results corroborate that our approach can more effectively comprehend contextual clues and generate more precise and natural answers, due to the integration of LLM. 

Additionally, we evaluate the VSR performance across various video length segments to explore the effectiveness of LLM in handling long speech. Figure \ref{fig:4} shows that WER decreases as video length increases. Notably, our proposed method exhibits outstanding recognition performance, with a WER of 12.9\% on videos longer than 6 seconds. Furthermore, our method demonstrates consistent performance improvements as the length of the video increases, compared to other methods. It indicates the effectiveness of LLM's context modeling in longer video utterances, which demand a more comprehensive understanding of context.

\begin{table}[t]
\renewcommand{\arraystretch}{1.6}
  \centering
  \resizebox{0.999\linewidth}{!}{
  \begin{tabular}{ccccccccc}
    \Xhline{3\arrayrulewidth}
    \multirow{2}{*}{\makecell{\textbf{Number of} \\ \textbf{Clusters}}} & \multicolumn{5}{c}{\textbf{BLEU  $\uparrow$}} & \multirow{2}{*}{\makecell{\textbf{Length of} \\ \textbf{sequence}}} & \multirow{2}{*}{\makecell{\textbf{FLOPs (P)}}} \\
    \cline{2-6}
    & \textbf{En-It} & \textbf{En-Fr} & \textbf{En-Pt} & \textbf{En-Es} & \textbf{Avg} & \\
    \hline 
    - & 12.3 & 15.8 & 13.7 & 16.7 & 14.6 & 1.00 & 62.4\\
    \cdashline{1-8}
    2000 & 11.2 & 15.9 & 13.8 & 16.5 & 14.4 & 0.70 & 53.8 (13.8\%) \\
    200 & 12.1 & 15.4 & 13.6 & 16.8 & 14.5 & 0.53 & 45.6 (26.9\%) \\
    50 & 12.1 & 14.9 & 13.3 & 16.9 & 14.3 & 0.45 & 41.0 (34.3\%) \\
    
    \Xhline{3\arrayrulewidth}
  \end{tabular}}
  \caption{Analysis on computational efficiency with varying number of visual speech unit clusters. When the deduplication strategy is adopted, the proposed method obtains comparable performances with greatly reduced sequence length and training FLOPs.}
  \vspace{-0.4cm}
  \label{table:3}
\end{table}


\begin{figure}[t]
	\centering
	\centerline{\includegraphics[width=0.999\linewidth]{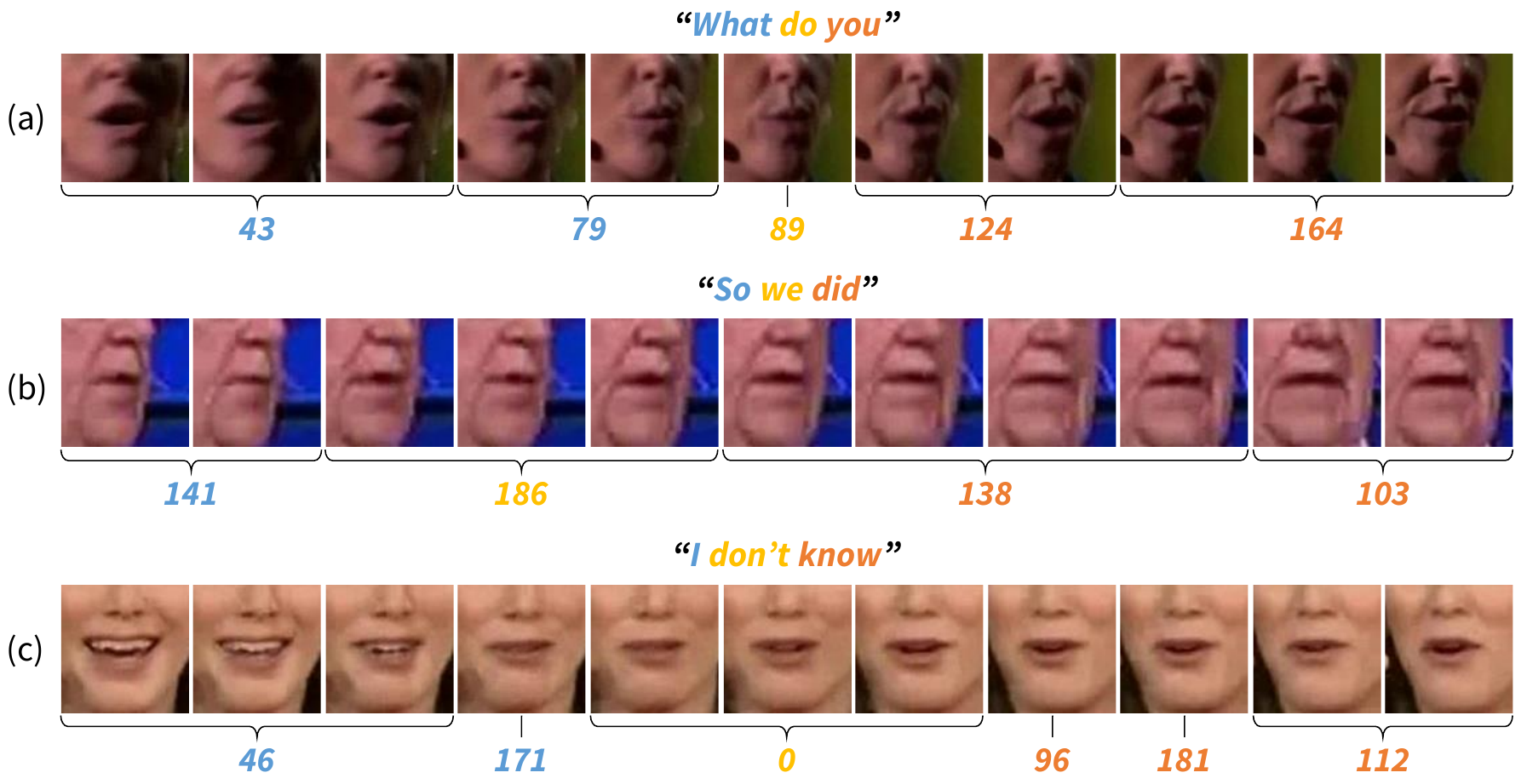}}
        \caption{Visualization results showing how video frame features are deduplicated and mapped into visual speech units. By doing so, the redundant frame features can be reduced efficiently.}
	\label{fig:5}
    
\end{figure}

\subsubsection{Effectiveness of Deduplication}
We conduct experiments to assess the effectiveness of our deduplication strategy. For the deduplication process, the number of clusters for visual speech units is required to be determined, and we show the effectiveness according to the number of clusters. Table \ref{table:3} presents these results, and the first row shows the performance of the baseline which does not utilize the deduplication. The baseline obtains an average BLEU score of 14.6 with 62.4 peta FLOPs per training epoch. By applying the proposed deduplication, our method acquires comparable performance, while significantly reducing the sequence length and computational resources (FLOPs). Specifically, with 200 clusters for visual speech units, our method not only maintains a similar performance level with a 14.5 average BLEU score but also cuts the sequence length by 53\%. Consequently, the FLOPs are greatly reduced to 45.6, marking a 26.9\% decrease. These experiments confirm that deduplication, applied to visual speech units, effectively eliminates redundant information.

Moreover, we delve into the deduplication process by examining it at the video frame level to check whether consecutive visual features, characterized by similar lip movements, are grouped into the same visual speech unit. Figure \ref{fig:5} provides several visual examples alongside their corresponding phrases and video frames. In Figure 
\ref{fig:5} (a), as a speaker articulates “\textit{What do you}”, it's noted that 11 video frames can be expressed by 5 visual speech units. For instance, the visual sequences for the sound “wha” belong to the same 43rd unit. Similarly, Figure \ref{fig:5} (c) illustrates that the four frames corresponding to “I” can be efficiently represented by the 46th and 171st visual speech units. Through this analysis, we confirm that visual features with similar lip shapes can be effectively deduplicated, significantly reducing the visual sequence's length.

\subsubsection{VSP-LLM in Data-limited Situation}
Leveraging the contextual understanding capabilities of LLM, which are pre-trained on vast text corpora, we suppose that a small amount of labeled data is sufficient for constructing a unified VSR and VST model. This is because the proposed VSP-LLM endeavors to establish visual-to-text mapping while entrusting the task of language modeling to the LLM. To validate it, we train VSP-LLM on the MuAViC dataset with different amounts of labeled data; \textbf{1 hour}, \textbf{5 hours}, \textbf{10 hours}, and \textbf{15 hours}. For comparison, we also develop AV-HuBERT on the same data. Table \ref{table:4} displays the VSR and VST performances. In all experimental conditions, regardless of the amount of data used, our proposed method significantly outperforms AV-HuBERT. Moreover,  when using only 15 hours of labeled data, our unified method achieves a WER of 32.8\%. This is a noteworthy achievement, particularly when compared to the previous VSR \cite{makino2019recurrent} model achieving a WER of 33.6\%, by using 31k hours of labeled data for training.

\begin{table}[t]
\renewcommand{\arraystretch}{1.6}
  \centering  
  \resizebox{0.999\linewidth}{!}{
  \begin{tabular}{ccccccccc}
    \Xhline{3\arrayrulewidth}
    \multirow{2}{*}{\textbf{Method}} & \multirow{2}{*}{\makecell{\textbf{Labeled} \\ \textbf{Data(hrs)}}} & \multicolumn{5}{c}{\textbf{BLEU  $\uparrow$}} & \multirow{2}{*}{\textbf{WER(\%) $\downarrow$}}  \\
    \cline{3-7}
    & & \textbf{En-It} & \textbf{En-Fr} & \textbf{En-Pt} & \textbf{En-Es} & \textbf{Avg} & \\
    \hline 
    \textbf{AV-HuBERT} & 1 & 0.0 & 0.0 & 0.1 & 0.1 & 0.5 & 100.2 \\
    \textbf{VSP-LLM}& 1 & 1.0 & 2.8 & 2.0 & 1.7 & 1.8 & 84.84  \\
    \cdashline{1-8}
    \textbf{AV-HuBERT} & 5 & 1.4 & 3.8 & 2.0 & 1.7 & 2.2 & 71.9 \\ 
    \textbf{VSP-LLM} & 5 & 10.6 & 14.0 & 11.5 & 15.1 & 12.8 & 36.2 \\
    \cdashline{1-8}
    \textbf{AV-HuBERT} & 10 & 3.0 & 5.1 & 3.9 & 4.5 & 4.1 & 56.7 \\ 
    \textbf{VSP-LLM} & 10 & 12.1 & 15.4 & 13.6 & 16
    8 & 12.8 & 34.3 \\
    \cdashline{1-8}
    \textbf{AV-HuBERT} & 15 & 3.4 & 7.1 & 5.5 & 8.7 & 6.2 & 52.4 \\ 
    \textbf{VSP-LLM} & 15 & 13.5 & 16.9 & 14.2 & 17.0 & 15.4 & 32.8 \\ 
    \Xhline{3\arrayrulewidth}
  \end{tabular}}
  
  \vspace{-0.2cm}
  \caption{Impact of the amount of labeled data. It shows that a small amount of labeled data is sufficient to construct a unified VSR and VST model by leveraging contextual understanding capabilities of LLM.}
  \label{table:4}
  \vspace{-0.5cm}
\end{table}

\label{sec:4.3.4}
\vspace{-0.1cm}
\section{Conclusion}
\vspace{-0.1cm}
In this paper, we proposed a novel framework, Visual Speech Processing with LLMs (VSP-LLM), designed to leverage the context modeling ability of LLMs. Through this framework, we built a unified model that can perform multi-tasks, VSR, and VST, with a single model. Moreover, the proposed deduplication strategy reduces the redundant information of visual speech representations based on pronunciation information modeled from visual speech units. Through extensive experiments, we verified that the proposed deduplication method can reduce the visual sequence length by about 50\% with minimal performance degradation. In addition, we validated the effectiveness of the VSP-LLM by achieving a superior performance in the MuAViC benchmark with only 30 hours of labeled data.

\section{Limitations}
We have proposed a powerful visual speech processing method that incorporates LLMs to recognize and translate lip movements into other languages, leveraging the rich context modeling ability of LLMs. Despite the impressive improvement in the performance of this proposed method, the utilization of LLMs has been limited to VSR and VST tasks. We expect that the proposed VSP-LLM framework can be expanded to in real-world communication scenarios by utilizing additional non-verbal cues such as facial expressions and gestures. Especially, the VSP-LLM combined with non-verbal cues is expected to perform various tasks such as emotional recognition and dialog generation, starting with this paper as a foundation.

\section{Broader impact and ethics}
The integration of Large Language Models (LLMs) within our framework plays a pivotal role in its ability to handle the complexities of visual speech across different languages. LLM brings a deep understanding of contextual and linguistic information, which is critical for accurately interpreting and translating visual speech cues. This capacity for nuanced language processing underpins our confidence in the framework's potential for broader linguistic applicability. Moreover, our experiments have demonstrated exceptional data efficiency and significant performance gains with relatively small amounts of labeled data for each language. This efficiency is crucial for scalability to other languages and dialects, particularly those for which extensive labeled datasets may not be readily available. The ability to achieve robust performance with limited data is indicative of the framework's adaptability and its potential for expansion to a wider linguistic range.


\bibliography{main}
\bibliographystyle{acl_natbib}

\newpage
\appendix

\clearpage
\label{sec:sup}

\begin{figure}[t]
	\centering
	\centerline{\includegraphics[width=0.999\linewidth]{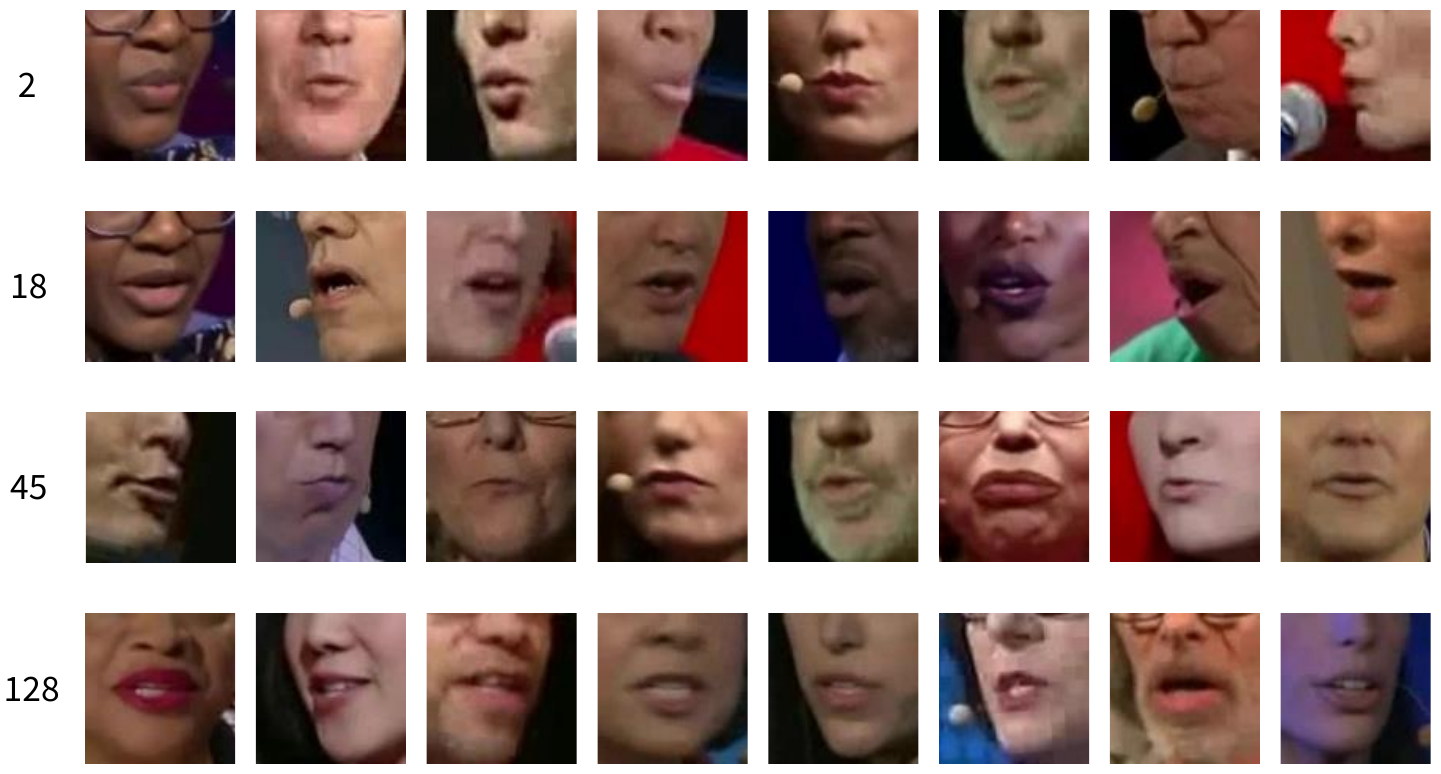}}
	\caption{Visualization of video frames corresponding to visual speech units. Each number indicates an index of visual speech unit.}
	\label{fig:6}
\end{figure}
\begin{table}[t]
\renewcommand{\arraystretch}{1}
\renewcommand{\tabcolsep}{7.0mm}
  \centering
  
  \resizebox{0.999\linewidth}{!}{
  \begin{tabular}{cc}
    \Xhline{3\arrayrulewidth}
    \textbf{Number of Clusters} & \textbf{FLOPs (P)} \\
    \hline
    w/o deduplication & 19.2 \\ 
    2000 & 16.2 (15.6\%) \\
    200 & 14.0 (27.1\%) \\ 
    50 & 12.6 (34.4\%) \\ 
    \Xhline{3\arrayrulewidth}
  \end{tabular}}

  \caption{Analysis on computational efficiency with varying number of visual speech unit clusters in inference time.}
  \vspace{-0.6cm}
  \label{table:5}
\end{table}

\section{Visualization of Visual Speech Units}
The visualization results of the visual speech units are shown in Figure \ref{fig:6}. In this paper, we use 200 clusters in order to generate visual speech units. Through analyzing the results, we verify that the video frames assigned the same visual speech unit have similar lip movement.

\section{FLOPs During Inference with Deduplication}
Table \ref{table:5} shows the FLOPs during inference time. Similar to during training, applying deduplication techniques also significantly reduced inference FLOPs.

\begin{table*}[t]
\renewcommand{\arraystretch}{1.5}
\renewcommand{\tabcolsep}{1mm}
  \centering
  
  \resizebox{0.999\linewidth}{!}{
  \begin{tabular}{l l}
    \Xhline{3\arrayrulewidth}
    \textbf{Sample ID} & \textbf{Label} \\
    \hline
    test/VIgzTLDyObo/00004 & \textbf{and then what happens} \\ 
    trainval/jpeSLKnS4gM/50020 & \textbf{and then what happens} \\
    \cdashline{1-2}
    test/vXPJVwwEmiM/00004 & \textbf{you probably won't} \\ 
    pretrain/omGbKQIzoWY/00009\_00 & \makecell{\textbf{you probably won't} do well on that problem on the other hand relaxed daydreaming is a way to} \\ 
    \Xhline{3\arrayrulewidth}
  \end{tabular}}

  \caption{Examples of cases where sentences in the test set also appear in the training set, but are spoken by distinct individuals.}
  \label{table:6}
\end{table*}
\section{Exposure to Transcriptions in the Pre-Training of LLM}
There might be concerns regarding LLaMA2's potential exposure to the LRS3 dataset during the pre-training phase. Since the details of LLaMA2's training data aren't publicly available, we can't be absolutely sure whether LRS3 was included or not. However, it's important to emphasize that the core challenge and focus of visual speech recognition (VSR) and translation (VST) lie in the ability to accurately match mouth shapes to unseen speakers, rather than merely replicating text from specific sentences. In particular, the mouth shape of the same sentence can vary significantly when expressed by different speakers, emphasizing the visual rather than textual nature of the work. Our analysis of the LRS3 dataset (Table \ref{table:6}) highlights this point, showing cases where sentences in the test set also appear in the training set, but are spoken by distinct individuals. This case serves to highlight the importance of the model's ability to recognize speaker-specific mouth shapes over memorizing textual content. Given this context, we believe that the potential exposure of LLaMA2 to certain sentences from the LRS3 dataset during training is unlikely to significantly impact the model’s performance in our study.

\begin{table*}[]
\renewcommand{\arraystretch}{1.5}
\renewcommand{\tabcolsep}{7mm}
\resizebox{0.999\linewidth}{!}{
\begin{tabular}{r:l}
\hline
    
\multicolumn{2}{c}{\textbf{Homophene Cases}}                                                       \\ \hline
\textbf{Ground   Truth } & it's not like teaching them how to \textcolor[rgb]{0.9,0,0}{ride} a bike \\
\textbf{\citet{prajwal2022sub} }      & it's all i teach them how to \textcolor[rgb]{0.9,0,0}{write} a bike    \\
\textbf{VSP-LLM   }      & it's not like teaching them how to \textcolor[rgb]{0.9,0,0}{ride} a bike   \\ \hline
\textbf{Ground   Truth } & is it about \textcolor[rgb]{0.9,0,0}{earning} as much as you possibly can   \\
\textbf{\citet{prajwal2022sub} }      & it's about \textcolor[rgb]{0.9,0,0}{learning} as much as possibly can    \\
\textbf{VSP-LLM   }      & it's about \textcolor[rgb]{0.9,0,0}{earning} as much as you possibly can    \\ \hline
\textbf{Ground   Truth } & it's like a piece of \textcolor[rgb]{0.9,0,0}{junk} mail to be thrown away \\
\textbf{\citet{ma2021end} }      & it's like a piece of \textcolor[rgb]{0.9,0,0}{chunk} made to be thrown away  \\
\textbf{VSP-LLM   }      & it's like a piece of \textcolor[rgb]{0.9,0,0}{junk} mail being thrown away  \\ \hline
\textbf{Ground   Truth } & and imagine what might happen because every \textcolor[rgb]{0.9,0,0}{region} has something to offer \\
\textbf{\citet{ma2021end} }      & and imagine what might happen because every \textcolor[rgb]{0.9,0,0}{reason} has something to offer \\
\textbf{VSP-LLM   }      & and imagine what might happen because every \textcolor[rgb]{0.9,0,0}{region} has something to offer \\
\hline
\end{tabular}}
\caption{Additional baseline examples for the homophene case. The \textcolor[rgb]{0.9,0,0}{Red} words indicate homophene words.}
\label{table:7}
\end{table*}
\section{Additional Examples of Homophene case}
In Section \ref{sec:rich_context}, we discussed the VSP-LLM model's exceptional ability to correctly distinguish homophenes by leveraging its advanced context modeling capabilities. This section further extends our analysis by comparing the performance of the VSP-LLM with other baseline models in handling homophenes. The results of these comparisons are presented in Table \ref{table:7}. In one notable example, Ma \etal incorrectly transcribed "junk" as "chunk." In contrast, the VSP-LLM accurately recognized the phrase "junk mail," a commonly used and contextually appropriate phrase in English. This illustrates the VSP-LLM's superior performance, particularly its proficiency in integrating contextual understanding with linguistic patterns to enhance transcription accuracy in cases involving homophenes.

\begin{figure*}[t]
	\centering
	\centerline{\includegraphics[width=16cm]{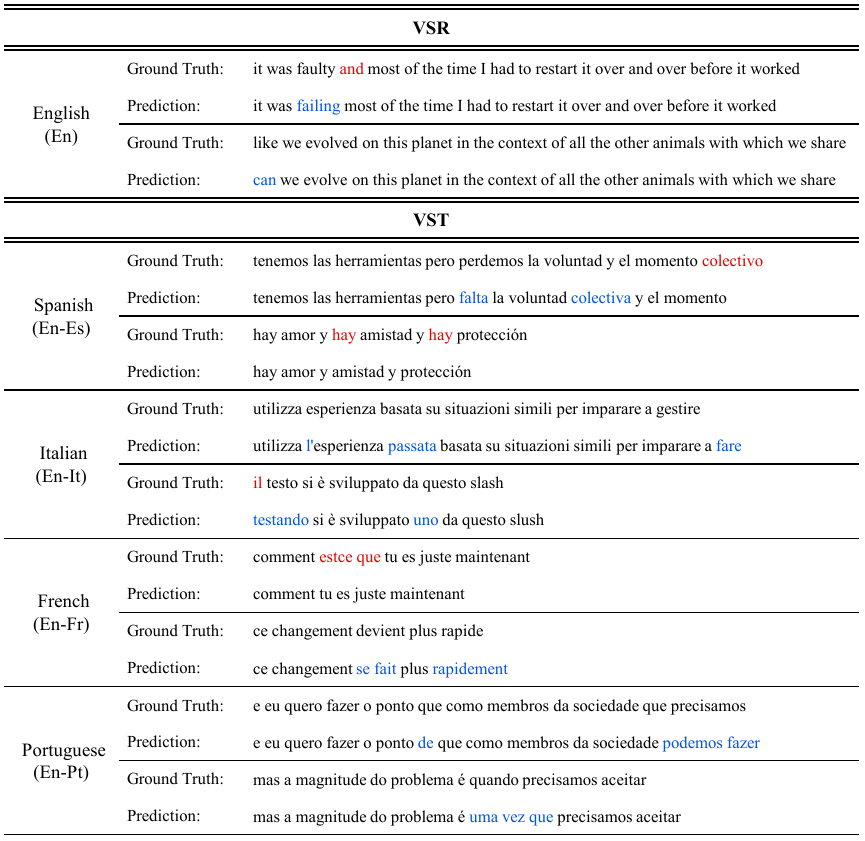}}
	\caption{Examples of VSR and VST predictions produced by our proposed model on LRS3 and En-to-X test set. Deletions from the ground-truth text are highlighted in \textcolor[rgb]{0.9,0,0}{Red}, while substitutions or addition are shown in  \textcolor[rgb]{0,0.2,0.90}{Blue}.}
	\label{fig:7}
\end{figure*}
\section{Examples of Predicted Sentences}
The examples of recognized and translated transcription by the proposed unified model are shown in Figure \ref{fig:7}. For generating transcription, we use a single-trained model that performs both VSR and VST tasks. 

\section{Erratum}
In Section 4 of this paper, we intended to report the translation results on the MuAViC dataset. However, the results were mistakenly reported on the LRS3-T dataset instead of the MuAViC dataset. In the revised manuscript, we have corrected this error by assessing and reporting the translation results on the correct MuAViC dataset.

\end{document}